\pdfoutput=1
\documentclass[11pt]{article}

\usepackage[final]{acl}

\usepackage{times}
\usepackage{latexsym}
\usepackage{multirow}
\usepackage{tabularx}
\usepackage{booktabs}
\usepackage{array}
\newcolumntype{L}[1]{>{\raggedright\arraybackslash}p{#1}}
\definecolor{lightyellow}{RGB}{255, 255, 204}
\usepackage[table]{xcolor}
\usepackage{booktabs}
\usepackage{makecell}
\usepackage{enumitem} 
\usepackage{caption}
\usepackage{booktabs}
\usepackage{siunitx}
\usepackage{array} % for >{\raggedright\arraybackslash}p{}
\usepackage{makecell} % for better line breaks inside table cells
\newcolumntype{C}[1]{>{\centering\arraybackslash}m{#1}}
\usepackage[table]{xcolor} % 放在导言区
\definecolor{lightgray}{gray}{0.95} % 自定义浅灰背景

\usepackage[T1]{fontenc}
\usepackage[utf8]{inputenc}

\usepackage{microtype}

\usepackage{inconsolata}
\usepackage{enumitem}
\usepackage{graphicx}
\definecolor{lightblue}{RGB}{16,98,180}
\definecolor{lightpink}{RGB}{243,40,109}
\definecolor{lightgreen}{RGB}{0,220,0}
\AtBeginDocument{%
  \hypersetup{%
    linkcolor=lightpink,
    urlcolor=lightpink,
    filecolor=lightblue
  }%
}

\title{ReEvalMed: Rethinking Medical Report Evaluation by Aligning Metrics with Real-World Clinical Judgment}

\author{Ruochen Li\textsuperscript{1*}, Jun Li\textsuperscript{1,2*}, Bailiang Jian\textsuperscript{1,2}, Kun Yuan\textsuperscript{1,4}, Youxiang Zhu\textsuperscript{3}\\
\textsuperscript{1}Technical University of Munich
\textsuperscript{2}Munich Center for Machine Learning \\
\textsuperscript{3}University of Massachusetts Boston
\textsuperscript{4}University of Strasbourg\\
 \small{
   {ruochen.li@tum.de, youxiang.zhu001@umb.edu}
 }
}

\begin{document}
\maketitle

\def\thefootnote{*}\footnotetext{Equal contribution.}\def\thefootnote{\arabic{footnote}}

\begin{abstract}
Automatically generated radiology reports often receive high scores from existing evaluation metrics but fail to earn clinicians’ trust. This gap reveals fundamental flaws in how current metrics assess the quality of generated reports. We rethink the design and evaluation of these metrics and propose a clinically grounded Meta-Evaluation framework. We define clinically grounded criteria spanning clinical alignment and key metric capabilities, including discrimination, robustness, and monotonicity. Using a fine-grained dataset of ground truth and rewritten report pairs annotated with error types, clinical significance labels, and explanations, we systematically evaluate existing metrics and reveal their limitations in interpreting clinical semantics, such as failing to distinguish clinically significant errors, over-penalizing harmless variations, and lacking consistency across error severity levels. Our framework offers guidance for building more clinically reliable evaluation methods. Project link is \url{https://ruochenli99.github.io/ReEvalMed/}
\end{abstract}

\section{Introduction}
Radiology reports constitute a fundamental component of clinical workflows, supporting diagnostic reasoning, treatment planning, and follow-up decisions~\citep{hager2024evaluation, vrdoljak2025review}. The continued advancement of vision-language models has enabled the direct generation of medical reports from imaging data~\citep{li2023llava, hartsock2024vision, chen2024vlm}. 
Although these generated reports often attain high scores on standard natural language processing (NLP) metrics, such as  BLEU~\citep{papineni2002bleu} and ROUGE-L~\citep{lin2004rouge}, their clinical adoption has been hindered by the lack of thorough clinical validity and reliability, which undermines clinician trust. 
This reflects a critical gap where the conventional NLP metric scores fail to align with real-world clinical utility~\citep{zhang2025gema}. High-scoring reports can still contain factual inaccuracies, logical inconsistencies, or omissions that could compromise patient safety and care~\citep{hartsock2024vision, wang2025survey}.

This discrepancy motivates a fundamental re-evaluation of how medical report generation is evaluated~\citep{jing2025reason, wang2024semantic}. Rather than relying exclusively on shallow matching or general language similarity, evaluation metrics should also interpret clinical semantics, such as clinically meaningful differences, and accurately reflect the potential impact of errors on patient care~\citep{gu2025radalign}. Moreover, as metrics are also used for training and benchmarking generative models, a metric that fails to capture what clinicians value may falsely incentivize unsafe outputs and thereby undermine confidence in these models. In Section~\ref{sec:rethink}, we analyze current LLM-based metrics and highlight key limitations in both their scoring design and evaluation methodology.

To address these issues, we propose a set of clinically grounded evaluation criteria, detailed in Section~\ref{sec:criteria}, that define what constitutes a clinically reliable evaluation metric. These criteria encompass two essential dimensions: (1) \textit{Alignment with clinical needs}, including accurate reporting of description, location, distance, and size; and (2) \textit{Core metric capabilities}, including discriminative ability, robustness to clinically insignificant variations, and monotonic sensitivity to increasing error severity.

Building upon these principles, we introduce a Meta-Evaluation framework in Section~\ref{sec:meta} with 12 evaluation aspects that serve as probes to assess whether a metric effectively captures clinical semantics. We construct a dataset of ground truth and rewritten reports (GT–ME pairs), annotated with clinical significance labels and explanations across diverse error types and evaluation aspects. Our experiments reveal the strengths and limitations of widely used metrics, offering actionable insights to guide the development of more reliable and clinically aligned evaluation methods.
Our contributions are as follows:

\begin{itemize}[itemsep=1pt, topsep=0pt, parsep=0pt, partopsep=0pt]
\item  \textbf{Rethinking LLM-based metrics.} Our detailed analysis of current LLM-based evaluation metrics reveals essential design flaws and limitations in their evaluation methodology.
% We analyze current LLM-based evaluation metrics and reveal their key design flaws and limitations in evaluating these metrics.

% \item  \textbf{Clinically grounded evaluation criteria}. We propose a set of clinically grounded criteria, developed in collaboration with clinicians and aligned with authoritative bodies such as the Fleischner Society~\citep{farjah2022fleischner}, ACR Lung-RADS~\citep{christensen2024lungrads}, and SCCT~\citep{leipsic2014scct}, to define what makes a good evaluation metric. These criteria support both metric assessment and future metric design.
\item \textbf{Clinically grounded evaluation criteria.} We propose a set of evaluation criteria co-developed with clinicians and aligned with established standards organizations such as the Fleischner Society~\citep{farjah2022fleischner}, ACR Lung-RADS~\citep{christensen2024lungrads}, and SCCT~\citep{leipsic2014scct}. These criteria provide a solid definition of a clinically meaningful metric, guiding both assessment and future metric design.

\item  \textbf{A unified Meta-Evaluation framework.} We introduce the first comprehensive Meta-Evaluation framework for medical report metrics. This framework enables a more rigorous evaluation by assessing a metric's alignment with clinical needs and its core capabilities.

% \item  \textbf{Empirical comparison of existing metrics}. We use our framework to benchmark existing metrics and investigate both their strengths and limitations and the reasons behind their observed behaviors.
\item \textbf{Empirical comparison of existing metrics.} Using our framework, we empirically benchmark widely used metrics. Our analysis reveals their strengths, limitations, and the underlying factors influencing their performance, offering insights for future development.
   
\end{itemize}

\section{Rethinking Clinical Report Evaluation: Limitations of Current Metrics} \label{sec:rethink}

\sisetup{
  table-format=2.2,
  detect-weight=true,
  detect-inline-weight=math
}

\begin{table*}[t]
\centering
\resizebox{0.95\textwidth}{!}{%
\begin{tabular}{l *{12}{S}}
\toprule
\textbf{Clinical relevance} & \multicolumn{6}{c}{\textbf{Significant}} & \multicolumn{6}{c}{\textbf{Insignificant}} \\
\cmidrule(lr){2-7} \cmidrule(lr){8-13}
\textbf{Error category} & \textbf{1} & \textbf{2} & \textbf{3} & \textbf{4} & \textbf{5} & \textbf{6}  
                       & \textbf{1} & \textbf{2} & \textbf{3} & \textbf{4} & \textbf{5} & \textbf{6} \\
\midrule
\textbf{BERTScore reports} & 0.540 & 0.451 & 0.380 & 0.398 & 0.258 & 0.308 & 0.163 & 0.253 & 0.321 & 0.313 & -0.044 & 0.270 \\ 
\textbf{BLEU reports}      & 0.553 & 0.414 & 0.337 & 0.242 & 0.387 & 0.263 & 0.200 & 0.209 & 0.129 & 0.280 & -0.034 & -0.032 \\
\textbf{Radgraph reports}  & 0.454 & 0.421 & 0.424 & 0.278 & 0.118 & 0.412 & 0.238 & 0.295 & -0.026 & -0.031 & -0.057 & 0.216 \\
\textbf{S-Emb reports}     & 0.321 & 0.443 & 0.227 & 0.297 & 0.434 & 0.124 & 0.199 & 0.210 & 0.072 & -0.028 & -0.002 & 0.128 \\
\bottomrule
\end{tabular}
}
\caption{Average pairwise Pearson correlation of significant and insignificant error counts between six radiologists in the ReXVal dataset. 
Radiologists were presented with a ground-truth report from MIMIC-CXR~\citep{johnson2019mimic} and a generated report retrieved by a metric (e.g., BERTScore). 
They only labeled the number of clinically significant and insignificant errors, without providing any explanation.}
\label{tab:corr}
\end{table*}

In clinical settings, radiology reports are essential tools for clinicians, underpinning diagnostic reasoning and guiding medical decision-making. Recent advances in vision-language models have led to the development of systems capable of generating radiology reports directly from medical images and contextual inputs, with notable examples including MAIRA-2~\citep{bannur2024maira} and LLM-CXR~\citep{lee2023llm}. 
% Although generated reports from these models often achieve strong performance on standard metrics such as BLEU~\citep{papineni2002bleu} and ROUGE-L~\citep{lin2004rouge}, high scores on these benchmarks do not guarantee that clinicians find the reports trustworthy or suitable for use in real-world medical decision-making.
Although generated reports often perform well on standard metrics such as BLEU~\citep{papineni2002bleu} and ROUGE-L~\citep{lin2004rouge}, these scores are not indicative of clinical reliability or utility in real-world decision-making.

This discrepancy stems from fundamental limitations inherent in existing evaluation metrics:
Conventional approaches, such as BLEU and ROUGE-L, which assess lexical overlap; RadGraph F1~\citep{jain2021radgraph}, which measures entity extraction and alignment; and CheXbert-based classifiers~\citep{smit2020chexbert}, which focus on predefined abnormalities, primarily depend on surface-level matching. These methods lack a deeper understanding of the clinical meaning of the report and struggle to handle the complexity of real-world clinical scenarios.
Recently, Large language model (LLM)-based metrics have demonstrated improvements over traditional surface-level matching approaches. Notably, metrics such as GREEN~\citep{ostmeier2024green}, GEMA Score~\citep{zhang2025gema}, MRScore~\citep{liu2024mrscore}, and ReFINE~\citep{liu2024er2score}, leverage the six error categories defined in the ReXVal dataset to construct structured scoring tables. These scoring schemes not only capture clinically relevant error types, but also integrate subjective aspects, including readability, grammaticality, and coherence, which are commonly employed during both model training and final score computation. However, these metrics still exhibit limitations, revealing a clear gap between their scoring criteria and real-world clinical needs.

% \textbf{Coarse-Grained Scoring Schemes}. Many LLM-based evaluation metrics rely on scoring tables whose design is largely based on the six error categories defined in the ReXVal dataset: 
% (a) false prediction of findings,
% (b) omission of findings,
% (c) incorrect location or position,
% (d) incorrect severity,
% (e) mention of a comparison not present in the reference, and
% (f) omission of a relevant comparison describing change from a previous study.

% While these categories are clinically reasonable, they remain coarse and incomplete. Numerous clinically relevant aspects, such as size or Distance, Uncertainty expression, Internal Contradictions, and Description, are not adequately addressed. More details are in Section~\ref{sec:criteria}.
\textbf{Coarse-grained scoring schemes.} Many LLM-based evaluation metrics rely on scoring tables derived from the six error categories defined in the ReXVal dataset: 
\textit{(a)} false prediction, 
\textit{(b)} omission, 
\textit{(c)} incorrect location or position, 
\textit{(d)} incorrect severity, 
\textit{(e)} mention of a comparison not present in referenced impression, and 
\textit{(f)} omission of comparison describing a change from previous study.

While these categories are clinically reasonable, they remain coarse and incomplete. Several clinically relevant aspects, such as size, distance, uncertainty expression, internal contradictions, and descriptive accuracy, are not adequately captured. See Section~\ref{sec:criteria} for details.

\textbf{Questionable evaluation methodology.} The clinical validity of many recent evaluation metrics is assessed by computing Pearson Correlations between the metric’s scores and clinician annotations from the ReXVal dataset~\citep{yu2023radiology}. 
Specifically, six radiologists annotated 50 ground-truth and generated report sets (each set comprising one ground-truth and four candidate reports), labeling the number of errors per report across six predefined categories. Each error was further classified as clinically significant or insignificant. Metrics are then validated by measuring the Pearson correlation between their predicted scores and the aggregated radiologist-annotated error counts, under the assumption that higher correlations indicate stronger alignment with human preferences.

However, this assumption is problematic. As shown in Table~\ref{tab:corr}, we computed the average pairwise Pearson correlations among the six annotators for each error category across candidate reports. The resulting inter-annotator correlations are notably not high, indicating a lack of consensus among experts regarding the number and significance of errors. Consequently, a high Pearson correlation with these radiologist annotations alone is insufficient to demonstrate the clinical robustness or practical reliability of a metric.

\textbf{Insufficient alignment with clinical needs.} In clinical practice, radiology reports serve as a foundation for diagnosis, treatment planning, and medical decision-making. Clinicians place a premium on factual accuracy across critical aspects and are particularly sensitive to major logical inconsistencies or clinically significant errors. Meanwhile, they are generally tolerant of minor deviations that do not affect patient care, such as anatomically irrelevant details or stylistic variations.

Although the ReXVal dataset distinguishes clinically significant from insignificant errors, it only provides final error counts per category as judged by six annotators. Crucially, it does not document the rationale for these judgments, i.e., why an error was considered significant or insignificant in a given case. This lack of transparency prevents follow-up metrics from learning or modeling the clinical reasoning process behind these annotations. As a result, scoring tables and metrics derived from ReXVal are limited in their ability to truly align with clinician decision-making criteria. They reflect annotation outcomes but not the underlying clinical logic, making them insufficient for capturing the nuanced judgment clinicians apply when evaluating generated report.

% \section{What Makes a Good Metric for Clinical Report Evaluation?} 
\section{What Defines a Good Metric for Clinical Report Evaluation?}
\label{sec:criteria}
% Through consultations with experienced clinicians, we identify two essential requirements for an effective evaluation metric.
Based on consultations with clinicians and established clinical guidelines, we identify two essential requirements for effective evaluation metrics. These reflect the practical priorities clinicians consider when interpreting radiology reports and serve as foundational principles for metric design.

\begin{table*}[h!]
\small
\centering
\resizebox{0.90\textwidth}{!}{%
\begin{tabularx}{\textwidth}{L{1.8cm} L{6.8cm} L{6.0cm}}
\toprule
\textbf{Aspect} & \textbf{Significant Error} & \textbf{Insignificant Error} \\
\midrule

\multirow{2}{=}{Location} 
& GT: Multiple chronic appearing left-sided rib fractures 
& GT: New left retrocardiac opacity \\
& ME: Multiple chronic appearing right rib fractures 
& ME: New opacity behind the heart on the left side \\
\addlinespace

\multirow{2}{=}{Severity} 
& GT: Heart is mildly enlarged. 
& GT: Severe cardiomegaly \\
& ME: Severely enlarged heart 
& ME: Moderate-to-severe cardiomegaly \\
\addlinespace

\multirow{2}{=}{Description} 
& GT: An irregular mass with spiculated margins 
& GT: Bibasilar patchy ill-defined opacities \\
& ME: A round, smooth mass 
& ME: Bibasilar faint and poorly marginated opacities \\
\addlinespace

\multirow{2}{=}{Negation} 
& GT: No evidence of pneumothorax 
& GT: No pleural effusion is seen \\
& ME: Pneumothorax is present 
& ME: There is no definite pleural effusion \\
\addlinespace

\multirow{2}{=}{Modality} 
& GT: Refer to prior CT torso for full descriptive details of esophageal abnormalities. 
& GT: Consider chest CT for further evaluation \\
& ME: Refer to prior abdominal ultrasound for details of esophageal abnormalities 
& ME: CT can be considered for further assessment \\
\addlinespace

\multirow{2}{=}{Size Distance} 
& GT: Irregularly marginated 3-cm mass in the lingula 
& GT: ET tube within 1 cm of the carina \\
& ME: Irregularly marginated 8-cm mass in the lingula 
& ME: ET tube within 0.9 cm of the carina \\
\addlinespace

\multirow{2}{=}{Comparison} 
& GT: Pulmonary edema has improved 
& GT: No interval change in pleural effusion \\
& ME: Pulmonary edema has worsened
& ME: Pleural effusion is essentially unchanged \\
\addlinespace

\multirow{2}{=}{Internal Contradiction} 
& GT: The lungs are clear. 
& GT: No current evidence of larger pleural effusions \\
& ME: The lungs are clear. There is consolidation in the right base 
& ME: No current evidence of larger pleural effusions. Minimal pleural effusions may exist \\
\addlinespace

\multirow{2}{=}{Uncertainty} 
& GT: Whether this is pneumonia is radiographically indeterminate. 
& GT: A possible infiltrate is suggested \\
& ME: Pneumonia exists 
& ME: An infiltrate is likely present \\
\addlinespace

\multirow{2}{=}{Terminology} 
& GT: A cavitary lesion, suggesting tuberculosis 
& GT: A 3-cm mass \\
& ME: A hole, suggesting infection 
& ME: A 3-cm lesion \\

\midrule
\addlinespace
\multirow{2}{=}{Noise} 
& GT: Irregularly marginated 3-cm mass in the lingula has grown
& GT: subtle opacity may represent atelectasis \\
& ME: 3-cm lingula margins has been growing irregularly
& ME: subtble opaciti may represent atelectasi \\
\addlinespace

\multirow{2}{=}{Stylistic Variation} 
& GT: Bilateral left greater than right pleural effusion 
& GT: lung... pulmonary edema... pleural effusions... \\
& ME: Fluid accumulation on both sides of the chest, more on the right
& ME: pleural effusions... lung...pulmonary edema...\\

\bottomrule
\end{tabularx}
}
% \caption{Criteria Table and Examples of clinically significant and insignificant errors across evaluation aspects, shown as GT and ME(Meta-Evaluation Rewrite) pairs.}
\caption{Examples of clinically significant and insignificant errors across evaluation aspects, shown as Ground Truth (GT) and Meta-Evaluation Rewrite (ME) pairs.}
\label{tab:metric-error-pairs}
\end{table*}

\subsection{Alignment with Clinical Needs}
Clinicians are the primary users of medical reports and rely on them for downstream decisions such as diagnosis and treatment planning. Accordingly, what matters most is the accuracy and clinical reliability of the content. A single clinically significant error, such as misstating a tumor as 4 cm instead of 8 cm, can lead to entirely different clinical actions and potentially cause serious medical harm. By contrast, clinicians are generally tolerant of clinically insignificant deviations, such as the inclusion of benign incidental findings or the use of alternative but semantically equivalent expressions, as long as these do not interfere with diagnostic reasoning or therapeutic decision-making. Motivated by these practical considerations, we define a set of clinically grounded evaluation criteria that reflect the aspects clinicians prioritize when judging report quality. They are listed in the leftmost column of the first ten rows of Criteria Table~\ref{tab:metric-error-pairs}.

\textbf{Location} refers to the precise anatomical site of a lesion or abnormality (e.g., "right upper lobe"). Accurate localization is critical, as different sites often imply different etiologies and treatment strategies. Incorrect location may lead to diagnostic errors or inappropriate interventions.

\textbf{Severity} describes the extent or seriousness of a finding (e.g., "mild," "severe"). Although partially subjective, severity assessments inform clinical urgency and therapeutic prioritization. Incorrect severity may mislead clinicians regarding the risk level of a condition.

\textbf{Description} captures the morphological characteristics of a lesion, such as its size, shape, and edge properties (e.g., "well-defined,""irregular"). These features are key to differential diagnosis and malignancy assessment, directly affecting downstream clinical decisions.

\textbf{Negation} explicitly indicates the absence of certain abnormalities (e.g., "no pneumothorax"). Accurate negation narrows the differential diagnosis. Errors or omissions in negation may lead to serious misdiagnoses.

\textbf{Modality awareness} assesses whether the report and its recommendations are appropriate given the imaging modality used (e.g., X-ray, CT). Each modality has different resolution capabilities and clinical applications; failure to account for modality limitations may result in inappropriate conclusions.

\textbf{Size and distance} includes quantitative measurements of lesions (e.g., "4.5 cm") and positional relationships (e.g., "catheter tip 2 cm above the carina"). Such information is crucial for tumor staging, disease assessment, and device placement. Misstatements may directly affect treatment decisions and prognosis.

\textbf{Comparison and progression} captures temporal changes by comparing current findings to prior imaging (e.g., "increased," "unchanged"). It helps evaluate disease progression and treatment response, and guides follow-up planning. Omission or misstatement may disrupt clinical continuity.

\textbf{Internal contradiction} identifies logical inconsistencies within the same report (e.g., "clear lungs" in one sentence and "left lung infiltrate" in another). It undermines the credibility of the report and can lead to clinical misjudgment or patient harm.

\textbf{Medical terminology usage} assesses whether medical terms are used accurately and clearly. While some stylistic variation is acceptable, reports should avoid too ambiguous or misleading language. Incorrect terminology can impede accurate interpretation and decision-making.

\textbf{Uncertainty expression} involves the use of hedging terms (e.g., "may," "possibly," "suspicious for") to reflect diagnostic uncertainty. Properly expressed uncertainty helps clinicians plan differential diagnoses and additional tests. Omitting or misrepresenting uncertainty may result in overconfident or incorrect decisions.

\begin{figure}[]
    \centerline{\includegraphics[width=0.47\textwidth]{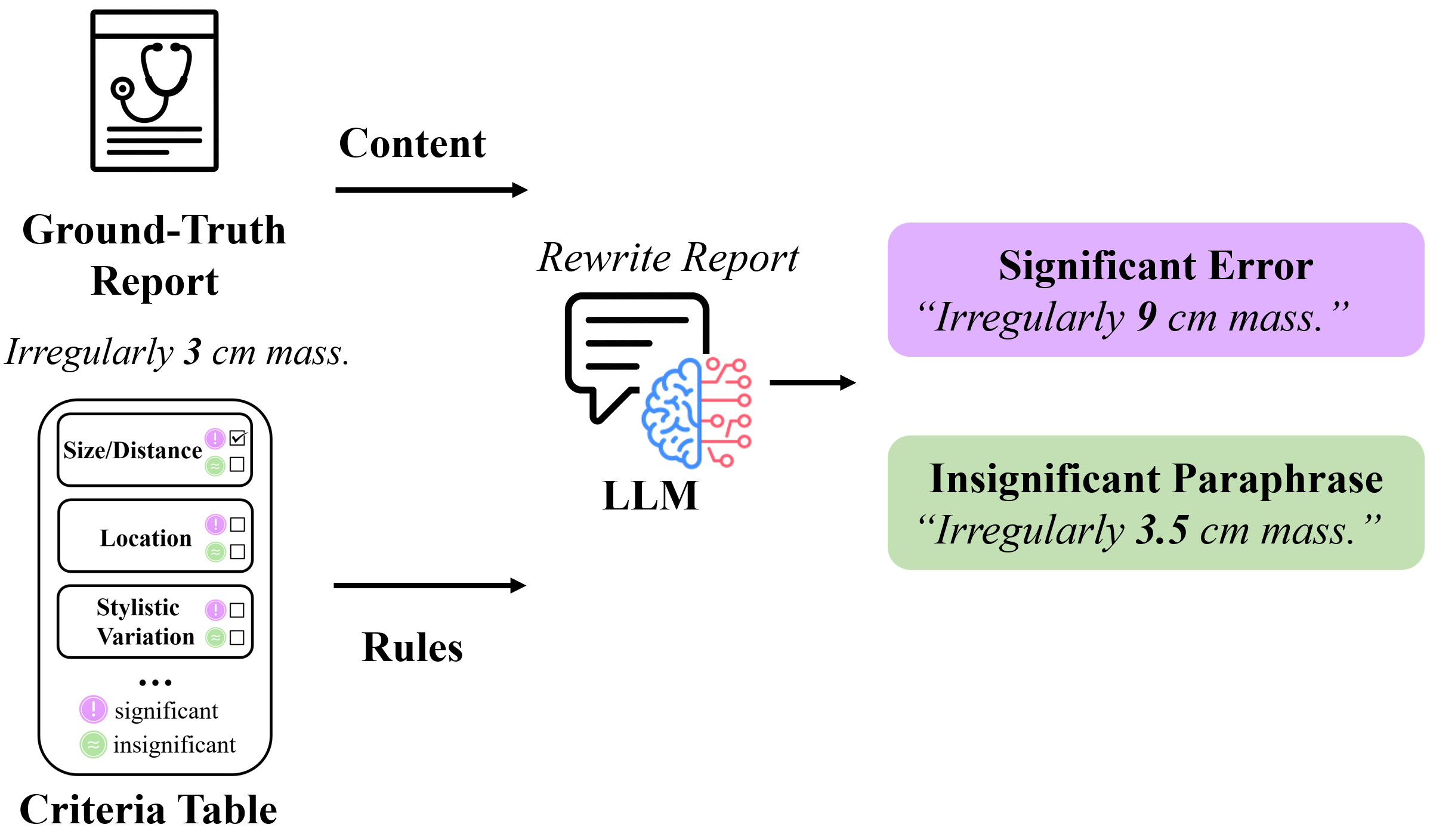}}
\caption{Rewrite report contains a significant error or an insignificant error.}
\label{fig:rewrite}
\end{figure}

\subsection{Metric Capabilities: Discrimination, Robustness, and Monotonicity}
Beyond alignment with clinical needs, a good metric should also possess the following capabilities:

\textbf{Discriminative ability.} Reliable metrics should be capable of distinguishing between clinically acceptable and clinically dangerous reports. Reports containing serious errors that could lead to adverse clinical decisions should receive substantially lower scores.

\textbf{Robustness ability.} Evaluation metrics should exhibit robustness to clinically insignificant variations, meaning they should avoid penalizing reports that differ from the reference only in superficial form, while remaining clinically equivalent in content. Examples are provided in Table~\ref{tab:metric-error-pairs}. Specifically, Grammatical noise refers to grammatical errors, typographical mistakes, or non-standard phrasing that do not affect the underlying clinical meaning.
Stylistic variation refers to differences in expression that do not alter the underlying clinical meaning (e.g., reordering sentences of findings). Such variation may arise from institutional templates, clinician-specific phrasing preferences, or differences in information ordering.

\textbf{Monotonicity}. A well-calibrated metric should exhibit a monotonic response to increasing clinical error severity, with scores consistently decreasing as errors become more serious. This property reflects the metric’s ability to not only detect errors, but to differentiate their clinical significance.

\section{Meta-Evaluation Framework: Dataset Construction and Metric Assessment} \label{sec:meta}

% \subsection{Dataset and Metric}
% We evaluate metrics using two primary sources of clinical data: the ReXVal dataset and the MIMIC-CXR dataset. From MIMIC-CXR~\citep{johnson2019mimic}, we randomly sampled 50 radiology reports. Additionally, we selected 20 information-rich reports from ReXVal, as identified by experienced radiologists with extensive clinical expertise.

% The evaluation includes the following reference-based metrics metrics from General NLP community:
% \textbf{BLEU}~\citep{papineni2002bleu}, \textbf{ROUGE-L}~\citep{lin2004rouge}, \textbf{METEOR}~\citep{banarjee2005}, and \textbf{BERTScore}~\citep{bert-score}.

% \textbf{AlignScore}~\citep{zha2023alignscore}, a factuality-based metric, assesses whether one sentence supports another (entailment). 

% \textbf{RaTEScore}~\citep{zhao2024ratescore}: a structured, entity-aware metric specifically designed for medical report evaluation. \textbf{CheXbert-F1}~\citep{smit2020chexbert} extracts 14 pre-defined thoracic disease labels and classifies their status. The final micro F1 score only evaluates exact label matches. \textbf{RadGraph-F1} ~\citep{jain2021radgraph} converts free-text reports into structured graphs via named entity recognition (NER) and relation extraction (RE). \textbf{GREEN}~\citep{ostmeier2024green}: a LLM-based metric that identifies and explains clinically significant errors in generated reports.

\subsection{Dataset and Metric}
We evaluate metrics using two primary sources of clinical data: the ReXVal dataset and the MIMIC-CXR dataset. From MIMIC-CXR~\citep{johnson2019mimic}, we randomly sampled 50 radiology reports. Additionally, we selected 20 information-rich reports from ReXVal, as identified by experienced radiologists with extensive clinical expertise.

The evaluation includes the following reference-based metrics from the general NLP community:  
BLEU~\citep{papineni2002bleu}, ROUGE-L~\citep{lin2004rouge}, METEOR~\citep{banarjee2005}, and BERTScore~\citep{bert-score}, which primarily assess surface-level lexical or semantic similarity. AlignScore~\citep{zha2023alignscore}, a factuality-based metric, assesses whether one sentence supports another (entailment). 

In addition to these general-purpose metrics, we include several domain-specific metrics tailored for medical report evaluation.  
RaTEScore~\citep{zhao2024ratescore} is a structured, entity-aware metric specifically designed for medical report evaluation.  
CheXbert-F1~\citep{smit2020chexbert} extracts 14 predefined thoracic disease labels and classifies their status. The final micro F1 score evaluates only exact label matches.  
RadGraph-F1~\citep{jain2021radgraph} converts free-text reports into structured graphs via named entity recognition (NER) and relation extraction (RE).  
GREEN~\citep{ostmeier2024green} is an LLM-based metric that identifies and explains clinically significant errors in generated reports.

\subsection{Discriminative Ability and Robustness}

% \begin{table*}[]
% \centering
% \resizebox{0.98\textwidth}{!}{%
% \begin{tabular}{C{1.0cm} L{17.0cm}}
% \toprule
% \textbf{Group} & \textbf{Example} \\
% \midrule
% 0 & "should be repositioned" $\rightarrow$ "consideration should be given to repositioning" \\
% 1 & "mild cardiomegaly" $\rightarrow$ "mild enlargement of the cardiac silhouette" \\

% \midrule
% 2 & "right lower and left upper lobes" $\rightarrow$ "left lower and right middle lobes"\\
% 3 & "No current evidence of pleural effusions, pulmonary edema, or pneumonia" $\rightarrow$ \newline
% \quad "consolidation in right middle lobe; moderate right-sided pleural effusion; small left apical pneumothorax" \\
% 4 & Report first states "The previous right internal jugular vein catheter was removed", then later fabricates "malpositioned right internal jugular catheter" \\
% \bottomrule
% \end{tabular}
% }
% \caption{Examples of rewritten errors across severity groups.}
% \label{tab:group_examples}
% \end{table*}

\begin{table*}[t]
\centering
\resizebox{0.9\textwidth}{!}{%
\begin{tabular}{c >{\raggedright\arraybackslash}p{15.5cm}}
\toprule
\textbf{Group} & \textbf{Example} \\
\midrule
0 & "should be repositioned" $\rightarrow$ "consideration should be given to repositioning" \\
1 & "mild cardiomegaly" $\rightarrow$ "mild enlargement of the cardiac silhouette" \\
\midrule
2 & "right lower and left upper lobes" $\rightarrow$ "left lower and right middle lobes" \\
3 & "No current evidence of pleural effusions, pulmonary edema, or pneumonia" 
     $\rightarrow$ "consolidation in right middle lobe; moderate right-sided pleural effusion; small left apical pneumothorax" \\
4 & Report first states "The previous right internal jugular vein catheter was removed", 
     then later fabricates "malpositioned right internal jugular catheter" \\
\bottomrule
\end{tabular}
}
\caption{Examples of rewritten errors grouped by severity, where a higher group ID indicates greater severity.}
\label{tab:group_examples}
\end{table*}

\begin{figure}[]
    \centerline{\includegraphics[width=0.47\textwidth]{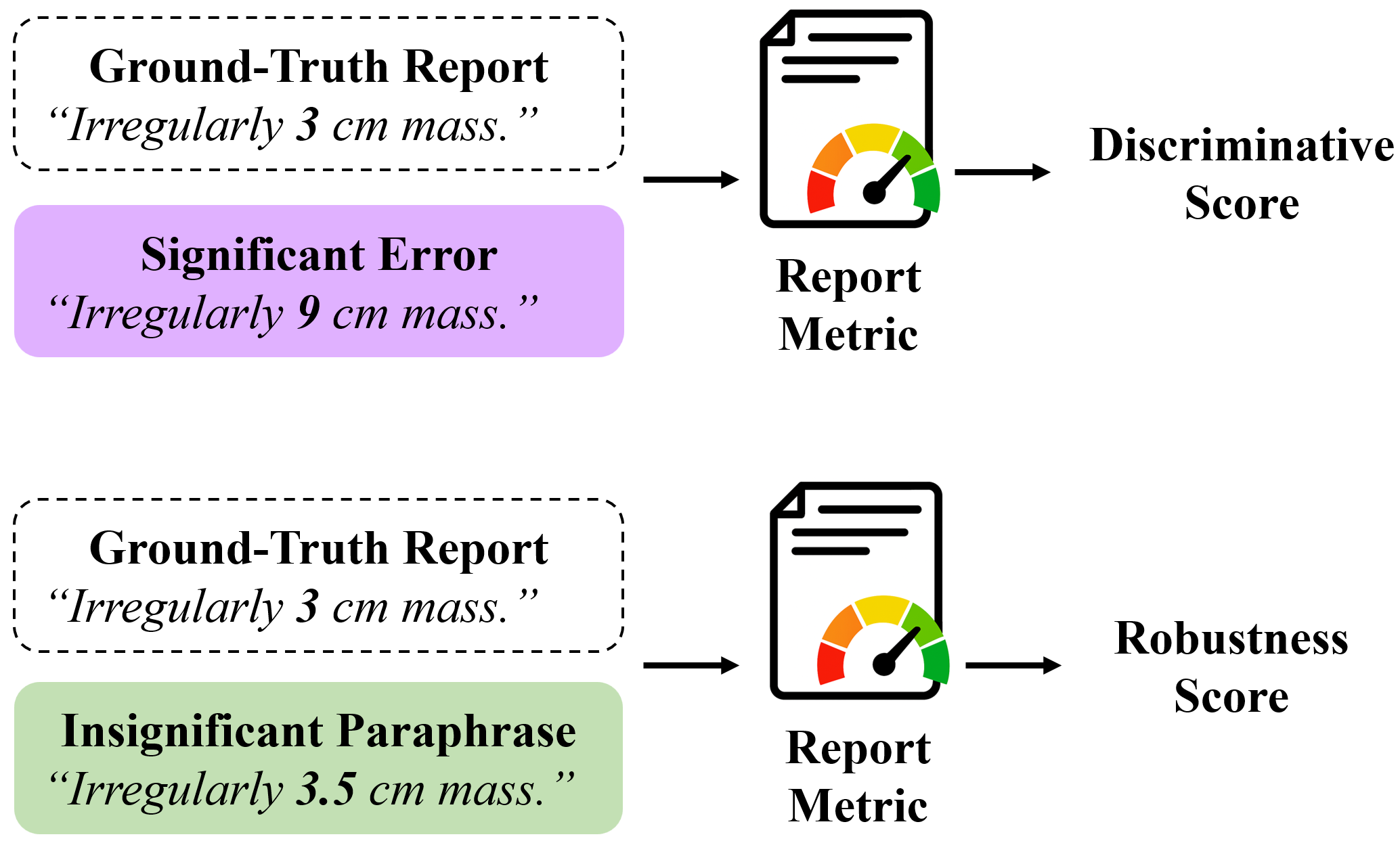}}
\caption{Discriminative Score and Robustness Score. }
\label{fig:score}
\end{figure}

\subsubsection{Rewrite Report}

We constructed a dataset of paired reports, each consisting of a Ground Truth (GT) report and a corresponding Meta-Evaluation Rewrite (ME), to test whether metrics can distinguish clinically significant from insignificant errors and remain robust to clinically irrelevant variations.

We categorized the differences between GT and ME into three primary error types.
\textbf{Omission}: A clinically relevant fact present in the GT is missing in the ME.
\textbf{Fabrication}: The ME introduces a clinically relevant fact that is not present in the GT.
\textbf{Inaccuracy}: The same clinical fact is described inconsistently between GT and ME.

For each evaluation aspect listed in the Criteria Table~\ref{tab:metric-error-pairs}, we prompted DeepSeek-R1~\citep{guo2025deepseekr1} to rewrite the GT report by modifying only one targeted aspect. The rewrites yield paired samples that contain either clinically significant or clinically insignificant errors.
For example, as shown in Figure~\ref{fig:rewrite}, under the Size/Distance aspect, an example clinically significant inaccuracy could be:

\textbf{GT:} \textit{"Irregularly marginated 3 cm mass."}

\textbf{ME:} \textit{"Irregularly marginated 9 cm mass."}

For those clinically high-impact aspects: Location, Severity, Description, and Comparison / Progression, we generated 10 significant and 10 insignificant error pairs for each of the three error types (omission, fabrication, inaccuracy). For the remaining aspects, we randomly sampled across error types while ensuring that each aspect contained 10 significant and 10 insignificant pairs. In total, the dataset comprises 400 expert-validated GT-ME report pairs. 

To preserve semantic fidelity and ensure isolation of the targeted error, we retained substantial contextual content in both GT and ME reports for significant error pairs, ensuring that the introduced change is the primary deviation. In contrast, for insignificant error pairs, we intentionally reduced the amount of surrounding context and kept only the modified part when appropriate, to prevent metrics from giving high scores just because the reports look similar overall.
All generated report pairs were reviewed and validated by experienced clinicians to ensure consistency with real-world clinical understanding and relevance, with explanations provided alongside each pair.

% We will publicly release the dataset, including both the original and rewritten reports, along with annotations indicating whether each modification constitutes a clinically significant or insignificant error, accompanied by brief justification.

\subsubsection{Discriminative and Robustness Score }

We applied multiple existing evaluation metrics to all constructed report pairs. For each metric, we separately computed the average scores for clinically significant and clinically insignificant error pairs, as shown in Figure~\ref{fig:score}.
The Discriminative Score is defined as the average score assigned to clinically significant error pairs. Lower scores indicate that the metric effectively penalizes critical errors, reflecting strong discriminative ability.
The Robustness Score is defined as the average score assigned to clinically insignificant error pairs. Higher scores suggest that the metric tolerates minor, clinically irrelevant variations, indicating desirable robustness.
Together, these two scores offer a comprehensive assessment of each metric’s capacity to distinguish between clinically meaningful and negligible errors. We further report confidence intervals to present a more transparent view of each metric’s consistency and variability.

\subsection{Test Monotonicity }

\begin{figure}[]
    \centerline{\includegraphics[width=0.47\textwidth]{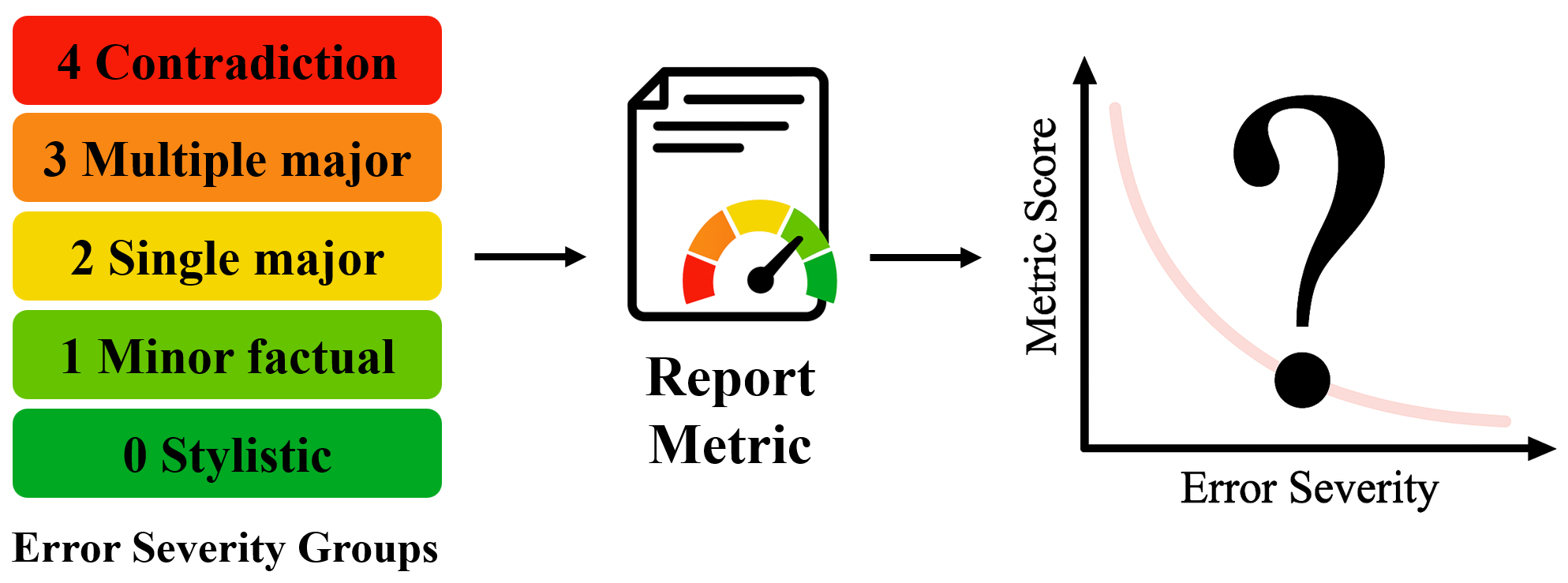}}
% \caption{ Testing Monotonicity.  }
\caption{Monotonicity evaluation using five error severity groups (Group 0–4), ranging from stylistic variations to severe logical contradictions.}

\label{fig:Error_sever}
\end{figure}

To assess whether evaluation metrics exhibit monotonic sensitivity to increasing error severity, we conducted a controlled test using a subset of the ReXVal dataset. We selected four GT reports and, for each, constructed corresponding ME reports with varying levels of error severity. These were organized into five groups (Group 0–4), each containing the same four GT–ME pairs, with severity increasing incrementally from minor stylistic variations to severe logical contradictions. Examples for each group are provided in Table~\ref{tab:group_examples}, and the overall grouping design is illustrated in Figure~\ref{fig:Error_sever}.

\textbf{Group 0: Stylistic Variation (Clinically Insignificant).} Contains purely stylistic or linguistic changes that do not alter clinical meaning. Examples include hedging expressions, synonymous reformulations, or reordering descriptive sentences.

\textbf{Group 1: Minor Factual Errors (Clinically Insignificant).} Involves minor factual inaccuracies that are clinically negligible and do not affect diagnostic interpretation or treatment decisions.

\textbf{Group 2: Single Error (Clinically Significant).} Introduces a single clinically significant error that may plausibly affect downstream clinical decisions.

\textbf{Group 3: Multiple Errors (Clinically Significant).} Introduces multiple (typically three) clinically significant errors, collectively increasing the risk of diagnostic misguidance.

\textbf{Group 4: Logical Contradiction (Severe Errors).} Introduces internal inconsistencies or logical contradictions, such as describing the presence of a structure previously stated to be absent. Such contradictions severely undermine clinician trust in the report and are considered critical failures.

For each group, we computed the average metric score across its four GT–ME pairs for each evaluation metric under consideration.
We then analyzed the score trajectory across severity levels to assess whether the metric exhibits a monotonic decreasing trend. A consistent decline in score as error severity increases indicates that the metric is well-calibrated. It appropriately reflects the clinical impact of errors through its scoring behavior.

\begin{figure}[]
    \centerline{\includegraphics[width=0.45\textwidth]{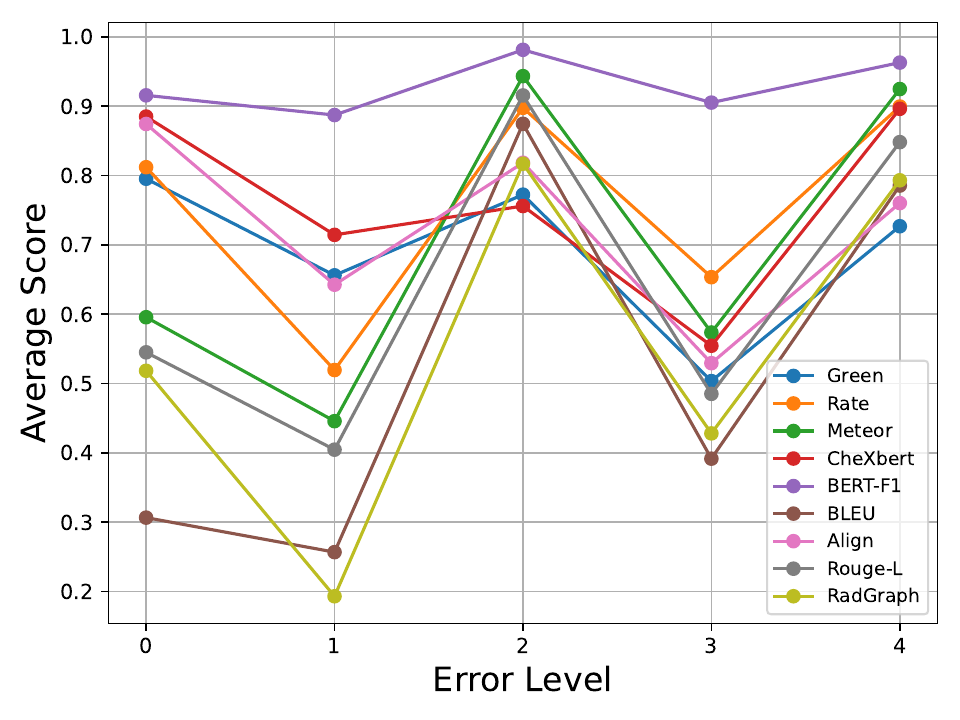}}
\caption{Metric scores vs. clinical error severity. Ideally, metric scores should decrease monotonically with increasing error severity.}
\label{fig:error_level}
\end{figure}

\begin{table*}[t]
\centering
\resizebox{0.99\textwidth}{!}{%
\begin{tabular}{l c c c| c |c c c | c c c | c| c| c | c c c |c| c | c |c| c |c}
\toprule
\textbf{Metric $\downarrow$} & \multicolumn{3}{c|}{\textbf{Comparison/Progression}} & \multicolumn{1}{c|}{\textbf{Cont.}} & \multicolumn{3}{c|}{\textbf{Description}} & \multicolumn{3}{c|}{\textbf{Location}} & \multicolumn{1}{c|}{\textbf{Mod.}} & \multicolumn{1}{c|}{\textbf{Neg.}} & \multicolumn{1}{c|}{\textbf{Noise}} & \multicolumn{3}{c|}{\textbf{Severity}} & \multicolumn{1}{c|}{\textbf{S/D}} & \multicolumn{1}{c|}{\textbf{S/V}} & \multicolumn{1}{c|}{\textbf{Unc.}} & \multicolumn{1}{c|}{\textbf{Term.}} & \textbf{Overall} & \textbf{CI}\\
 & \textbf{E} & \textbf{F} & \textbf{O} & \textbf{S} & \textbf{E} & \textbf{F} & \textbf{O} & \textbf{E} & \textbf{F} & \textbf{O} & \textbf{S} & \textbf{S} & \textbf{S} & \textbf{E} & \textbf{F} & \textbf{O} & \textbf{S} & \textbf{S} & \textbf{S} & \textbf{S} &  &  \\
\midrule
\rowcolors{2}{lightgray}{white}
Align & \textbf{61.03} & \textbf{71.54} & 88.81 & \textbf{65.02} & \textbf{59.92} & \textbf{73.91} & 93.43 & \textbf{50.94} & \textbf{52.66} & 95.01 & \textbf{55.40} & 79.22 & 74.96 & \textbf{71.17} & \textbf{80.88} & 96.58 & \textbf{77.60} & \textbf{49.94} & 72.52 & \textbf{59.38} & \textbf{71.50} & $[\mathbf{65.06}, \mathbf{77.93}]$\\
BERT-F1 & 98.40 & 97.54 & 96.55 & 96.03 & 97.56 & 98.29 & 97.35 & 99.46 & 98.41 & 97.27 & 95.06 & 97.72 & 94.89 & 99.19 & 98.39 & 97.97 & 99.12 & 93.47 & 97.73 & 95.90 & 97.31 & $[96.63, 98.00]$ \\
BLEU & 86.23 & 81.10 & \textbf{67.78} & 76.01 & 83.44 & 87.99 & 79.28 & 87.95 & 81.19 & \textbf{71.34} & 68.11 & 82.99 & \textbf{60.81} & 90.02 & 85.53 & 80.26 & 91.60 & 58.80 & 75.53 & 73.63 & 78.48 & $[74.37, 82.59]$ \\
CheXbert & 96.67 & 94.10 & 96.23 & 85.62 & 80.71 & 93.87 & 89.96 & 99.89 & 93.94 & 93.44 & 80.89 & \textbf{69.22} & 85.82 & 95.44 & 95.27 & 99.27 & 94.38 & 79.11 & \textbf{60.27} & 69.42 & 87.68 & $[82.79, 92.56]$ \\
GREEN & 67.89 & 77.90 & 78.44 & 68.08 & 69.45 & 83.14 & \textbf{77.88} & 66.55 & 74.57 & 74.90 & 69.17 & 80.55 & 73.98 & 73.98 & 82.73 & \textbf{74.67} & 83.87 & 51.15 & 83.81 & 73.21 & 74.30 & $[70.90, 77.69]$ \\
METEOR & 94.17 & 93.47 & 78.95 & 86.24 & 90.16 & 97.92 & 86.36 & 96.32 & 97.88 & 84.78 & 79.86 & 90.59 & 81.28 & 95.33 & 98.15 & 91.03 & 97.22 & 73.11 & 89.34 & 85.82 & 89.40 & $[86.23, 92.57]$ \\
RadGraph & 85.66 & 86.06 & 85.63 & 78.57 & 77.03 & 86.07 & 79.79 & 90.47 & 84.51 & 79.61 & 72.60 & 89.24 & 68.72 & 87.00 & 88.40 & 87.96 & 92.01 & 58.22 & 86.33 & 63.08 & 81.35 & $[77.28, 85.42]$ \\
RaTE & 89.49 & 89.42 & 84.74 & 87.66 & 89.96 & 95.98 & 90.48 & 93.11 & 95.68 & 84.69 & 88.15 & 91.69 & 79.76 & 90.38 & 95.29 & 91.02 & 96.78 & 78.43 & 92.67 & 78.64 & 89.20 & $[86.77, 91.63]$ \\
Rouge-L & 91.32 & 90.14 & 86.10 & 83.31 & 89.75 & 93.94 & 91.10 & 94.23 & 91.46 & 89.07 & 77.17 & 88.67 & 75.44 & 94.51 & 93.01 & 93.33 & 95.84 & 69.02 & 88.18 & 82.39 & 87.90 & $[84.77, 91.02]$ \\
\bottomrule
\end{tabular}
}
\caption{Discriminative Score (clinically significant errors). Lower values indicate better discrimination of errors. \textbf{Cont.}: Contradiction. \textbf{Mod.}: Modality. \textbf{Neg.}: Negation. \textbf{S/D}: Size/distance. \textbf{S/V}: Stylistic Variation. \textbf{Unc.}: Uncertainty. \textbf{Term.}: Terminology. \textbf{E}: Inaccuracy error. \textbf{F}: Fabrication error. \textbf{O}: Omission error. \textbf{S}: Inaccuracy, fabrication, and omission errors are randomly and evenly distributed.}
\label{tab:correctness}
\end{table*}

\begin{table*}[t]
\centering
\resizebox{0.99\textwidth}{!}{%
% \rowcolors{2}{lightgray}{white} % 
\begin{tabular}{l c c c| c |c c c |c c c | c| c| c| c c c |c| c| c| c| c |c}
\toprule
\textbf{Metric $\uparrow$} & \multicolumn{3}{c|}{\textbf{Comparison/Progression}} & \multicolumn{1}{c|}{\textbf{Cont.}} & \multicolumn{3}{c|}{\textbf{Description}} & \multicolumn{3}{c|}{\textbf{Location}} & \multicolumn{1}{c|}{\textbf{Mod.}} & \multicolumn{1}{c|}{\textbf{Neg.}} & \multicolumn{1}{c|}{\textbf{Noise}} & \multicolumn{3}{c|}{\textbf{Severity}} & \multicolumn{1}{c|}{\textbf{S/D}} & \multicolumn{1}{c|}{\textbf{S/V}} & \multicolumn{1}{c|}{\textbf{Unc.}} & \multicolumn{1}{c|}{\textbf{Term.}} & \textbf{Overall} & \textbf{CI} \\
 & \textbf{E} & \textbf{F} & \textbf{O} & \textbf{S} & \textbf{E} & \textbf{F} & \textbf{O} & \textbf{E} & \textbf{F} & \textbf{O} & \textbf{S} & \textbf{S} & \textbf{S} & \textbf{E} & \textbf{F} & \textbf{O} & \textbf{S} & \textbf{S} & \textbf{S} & \textbf{S} &  &  \\
\midrule
Align & 52.22 & 38.03 & 76.92 & 51.59 & 36.60 & 27.00 & 94.29 & 88.71 & 51.73 & 83.13 & 63.45 & 61.08 & \textbf{95.70} & 51.77 & 57.55 & 83.52 & 25.82 & 74.77 & 80.52 & 66.73 & 63.06 & $[53.67, 72.45]$ \\
BERT-F1 & 91.25 & \textbf{93.83} & 93.57 & \textbf{95.18} & \textbf{89.08} & \textbf{95.18} & \textbf{95.61} & 94.53 & \textbf{95.14} & 97.35 & 88.66 & \textbf{88.91} & 92.99 & 90.02 & \textbf{93.36} & 97.55 & 93.81 & \textbf{89.32} & \textbf{94.19} & \textbf{90.21} & \textbf{92.99} & $[\mathbf{91.75}, \mathbf{94.22}]$ \\
BLEU & 29.76 & 52.95 & 37.36 & 56.30 & 23.91 & 57.79 & 58.05 & 57.20 & 58.46 & 61.89 & 21.99 & 24.90 & 40.55 & 0.00 & 45.28 & 56.27 & 49.11 & 29.81 & 45.45 & 30.74 & 41.89 & $[34.61, 49.16]$ \\
CheXbert & \textbf{94.93} & 92.36 & \textbf{97.48} & 84.57 & 83.06 & 91.40 & 90.02 & \textbf{95.35} & 88.83 & \textbf{99.25} & 86.55 & 83.83 & 93.61 & \textbf{94.61} & 88.64 & \textbf{98.93} & \textbf{94.95} & 80.19 & 89.68 & 80.03 & 90.41 & $[87.82, 93.00]$ \\
GREEN & 81.67 & 80.00 & 76.67 & 67.17 & 75.00 & 63.33 & 75.00 & 78.24 & 66.33 & 73.33 & \textbf{91.00} & 82.50 & 84.17 & 60.00 & 56.67 & 70.00 & 90.00 & 65.05 & 77.50 & 81.67 & 74.76 & $[70.60, 78.93]$ \\
METEOR & 60.02 & 88.01 & 71.33 & 84.39 & 53.87 & 86.68 & 73.48 & 79.94 & 92.30 & 82.22 & 44.03 & 49.20 & 70.98 & 41.61 & 83.43 & 81.62 & 81.99 & 50.28 & 77.57 & 51.13 & 70.20 & $[63.02, 77.39]$ \\
RadGraph & 57.11 & 72.47 & 67.83 & 73.18 & 25.94 & 63.94 & 66.82 & 61.34 & 67.27 & 78.55 & 44.34 & 39.08 & 52.50 & 44.07 & 56.08 & 74.91 & 64.33 & 43.80 & 83.46 & 34.16 & 58.56 & $[51.66, 65.46]$ \\
RaTE & 64.30 & 84.60 & 71.32 & 83.57 & 63.11 & 85.55 & 89.81 & 78.03 & 82.62 & 84.02 & 63.25 & 64.57 & 75.49 & 75.15 & 79.71 & 80.50 & 78.13 & 72.26 & 88.58 & 63.80 & 76.42 & $[72.52, 80.31]$ \\
Rouge-L & 54.47 & 73.03 & 78.89 & 74.44 & 46.94 & 77.59 & 82.83 & 71.92 & 76.38 & 86.66 & 42.55 & 46.60 & 64.72 & 61.61 & 69.22 & 87.32 & 71.47 & 35.54 & 67.98 & 48.91 & 65.95 & $[59.26, 72.65]$ \\
\bottomrule
\end{tabular}
}
\caption{Robustness Score (clinically insignificant errors). Higher values indicate greater robustness to clinically irrelevant variations. Metrics with bold indicate top performance in that column.}
\label{tab:robustness}
\end{table*}

\section{Results and Analysis}
\subsection{Discriminative Ability and Robustness} 
We assess each metric’s ability to differentiate clinically significant and insignificant errors using the Discriminative Score and Robustness Score, as summarized in Tables~\ref{tab:correctness} and~\ref{tab:robustness}.

\textbf{General NLP metrics} lack clinical knowledge and aspect sensitivity. BLEU, METEOR, and ROUGE-L all exhibit poor robustness, frequently penalizing stylistic or structural variations that are clinically harmless. This stems from their reference-based design, which focuses on surface-level lexical or token overlap. BLEU shows relatively stronger discriminative ability, particularly for omission-related errors, but its extreme sensitivity to surface overlap leads to unjustified penalties. METEOR and ROUGE-L perform poorly on both scores, indicating limited clinical applicability. Overall, these metrics lack domain-specific understanding and fail to distinguish clinically significant from insignificant errors.

BERT-F1 achieves a high Robustness Score, indicating strong tolerance to clinically insignificant variations. However, it also yields a relatively high Discriminative Score, suggesting that it fails to adequately penalize clinically significant errors. This implies that while BERT-F1 is resistant to superficial changes, it lacks the sensitivity needed to distinguish harmful clinical deviations.

AlignScore, a factuality-based metric, has the lowest overall Discriminative Score as well as a low Robustness Score, indicating that it cannot reliably distinguish clinically significant from insignificant errors. This stems from its lack of medical knowledge, which limits its capacity to capture clinically relevant relationships, particularly in the presence of subtle but significant semantic shifts.

% BLEU, METEOR, and ROUGE-L all exhibit poor robustness, frequently penalizing stylistic or structural variations that are clinically harmless. While BLEU shows relatively better discriminative ability—particularly for omission-related errors—its extreme sensitivity to surface overlap causes unjustified penalties. METEOR and ROUGE-L perform poorly on both scores, indicating limited clinical applicability overall.

% RaTE score, although structured and entity-based, lacks sufficient medical knowledge grounding. It does not capture clinical priority or severity distinctions in a nuanced way. It shows reasonably high robustness but surprisingly yields an even higher Discriminative Score than its Robustness Score. This reflects its overreliance on surface-level entity matching and its inability to capture deeper clinical semantics or error impact.

\textbf{Non-LLM medical-specific metrics} often consider too few aspects or rely on rigid matching. RaTE, although structured and medical-entity-based, lacks sufficient medical grounding and fails to capture clinical priorities or severity distinctions in a nuanced way. While it demonstrates reasonably high robustness, its Discriminative Score is unexpectedly higher than its Robustness Score, reflecting an overreliance on surface-level entity matching and an inability to capture deeper clinical semantics or error impact. CheXbert-F1 gave consistently high scores to both types of errors, failing to reflect severity. This is because its final F1 score evaluates only exact matches against 14 predefined thoracic disease labels, ignoring semantic variations, contextual cues, and clinically equivalent paraphrases. 
RadGraph-F1, by contrast, shows a high Discriminative Score but a low Robustness Score. While theoretically powerful, it is overly sensitive to entity boundaries, relation formats, and exact phrasing. Even semantically equivalent rewrites, such as reordering or lexical variation, may reduce the score due to graph-matching failures, which highlights its poor robustness to stylistic variation and clinical equivalence.

\textbf{LLM-based medical-specific metrics} often rely heavily on ReXVal error categories, with a scoring framework that covers only a limited set of error types and lacks comprehensive coverage of clinically relevant semantic dimensions. This limitation extends to several newer LLM-based metrics that inherit similarly simplified scoring tables. GREEN achieves a better Discriminative Score than most metrics, performing particularly well on omissions of description and severity, indicating its ability to penalize clinically significant errors. However, it still suffers from a notably low Robustness Score, suggesting a tendency to over-penalize minor, clinically irrelevant differences and thus limiting its practical reliability.

% Overall, the results suggest that while some metrics perform well in either discrimination or robustness, none achieve both simultaneously. More importantly, many existing metrics lack the ability to capture clinical semantics, a capability that is essential for evaluating reports as a clinician would. These findings highlight the need for clinically grounded, error-aware evaluation approaches.
Overall, the results show that while some metrics perform well in either discrimination or robustness, none excel at both. Moreover, most existing metrics do not capture clinical semantics, which is essential to evaluate reports from the perspective of a clinician. These findings highlight the need for clinically grounded, error-aware evaluation.

\subsection{Monotonicity}
As illustrated in Figure~\ref{fig:error_level}, although different metrics exhibit varying absolute scores, their overall trends across severity groups are remarkably consistent.
From Group 0 to Group 1, all metrics show a decreasing trend, indicating that each metric can reliably distinguish stylistic variations (clinically negligible) from minor factual errors (still clinically insignificant but of slightly higher concern).
Similarly, from Group 2 to Group 3, we observe a consistent decline across all metrics, suggesting that they are sensitive to the increased number of clinically significant errors, even if this sensitivity may stem more from the extent of textual changes than from a true understanding of clinical severity.
These patterns indicate that existing metrics possess some limited ability to capture differences in error severity, particularly when differences are accompanied by large textual modifications.

However, two key transitions reveal important weaknesses.
From Group 1 to Group 2, all metrics unexpectedly show an increase in scores, implying a failure to distinguish clinically significant single errors from clinically insignificant factual deviations. This reversal can be attributed to our data set design: insignificant errors in Group 1 often involve larger surface-level changes (which do not affect clinical interpretation), while significant errors in Group 2 are more localized, with most of the surrounding context preserved, potentially misleading surface-based metrics.
From Group 3 to Group 4, metrics again show an increase in scores, reflecting their difficulty in detecting logical contradictions. This is likely because logical contradictions in Group 4 are introduced via small, localized insertions (e.g., contradicting earlier statements with a single sentence), while Group 3 reports contain multiple significant edits across the text. Metrics that rely heavily on overall textual similarity may struggle to penalize these subtle but clinically critical inconsistencies.

In summary, although existing metrics are sensitive to gross differences in error severity, they struggle with fine-grained distinctions, particularly in separating clinically significant from insignificant errors and in detecting logical contradictions.

\section{Conclusion}

In this paper, we rethink the design and evaluation of the existing metrics for medical report generation, arguing that effective metrics evaluation should not rely solely on coarse radiologists' counting-based annotations. 

To address this, we introduce a clinically grounded Meta-Evaluation framework and show that many existing metrics fail to capture clinical semantics, a critical requirement for evaluations that align with clinical judgement. 
To ground our Meta-Evaluation framework, we define clinical semantics as the fine-grained, decision-informing criteria within medical reports. We create an expert-annotated dataset of GT-ME report pairs to simulate a broad spectrum of clinically relevant scenarios and real-world diagnostic needs.
% Specifically, we define clinical semantics as the fine-grained criteria in medical reports that directly inform clinical decision-making in evaluating generated reports, and we construct an expert-annotated dataset of GT–ME report pairs that simulate a wide range of clinically relevant scenarios and real-world diagnostic needs. 
Our multi-dimensional framework enables a rigorous evaluation of clinical alignment and core metric capabilities, including a metric's discriminative ability, its robustness to clinically insignificant variations, and its monotonic sensitivity to increasing error severity.
Our findings uncover a critical misalignment between existing metrics and clinical needs.

General NLP metrics lack both clinical knowledge and aspect sensitivity. Likewise, existing medical-specific metrics often suffer from insufficient aspects and rigid matching. Although these metrics reflect thoughtful and valuable design efforts, our in-depth Meta-Evaluation concludes that the limitations of these metrics stem from the omission of clinically critical aspects during their formulation. We view these metrics as a strong foundation for further improvement and encourage future research to incorporate clinical aspects more explicitly. Our Meta-Evaluation framework can function as a diagnostic lens to pinpoint where current metrics fall short and serve as an important stepping stone toward developing more clinically aligned evaluation tools.

% our in-depth meta-evaluation reveals the fundamental flaws in such metrics design that omit the clinically critical aspects

\section*{Limitations}

% Scalability of Dataset Construction. Our current pipeline for constructing the meta-evaluation dataset requires substantial manual verification to ensure clinical accuracy, which limits its scale compared to datasets in other medical domains~\citep{sun2016skin, sun2024alice}. A more robust and scalable framework will be needed to support broader evaluations in the future.
% Limited Evaluation of Metric Interpretability. Interpretability and error localization are critical for clinical users to understand and trust evaluation metrics. While our criteria table implicitly supports fine-grained error attribution, we do not explicitly evaluate the interpretability of metrics in this work. Future studies will aim to assess this capability more directly.
\textbf{Scalability of dataset construction.} The current pipeline for building the Meta-Evaluation dataset requires manual verification to ensure clinical accuracy, limiting scalability compared to other medical domains~\citep{sun2016skin, sun2024alice}. A more automated and scalable framework is needed.

\textbf{Limited evaluation of metric interpretability.} Interpretability and error localization are essential for clinical users to understand and trust evaluation metrics. Although our criteria table supports fine-grained error attribution, we do not directly assess the metric's interpretability in this study. Future work will explore this dimension more explicitly.

\textbf{Limited Evidence for Modality Transferability.} Our criteria and Meta-Evaluation framework are designed to be modality-agnostic and theoretically generalizable to other domains (e.g., CT, MRI), but we have not validated them beyond CXR. Future work will explore broader imaging settings.

\bibliography{custom}

\newpage
\appendix

\section{Appendix: Aspect Explanation}
\label{sec:appendix a}
These aspect-level scores enable clinicians and researchers to assess whether existing metrics truly capture clinical semantics. Our dataset contains the explanation for each GT-ME pair.

\textbf{Location errors} can directly affect treatment decisions, especially in surgery or local therapy. Misreporting lesion sites may lead to incorrect staging, inappropriate treatment, or unnecessary procedures. Significant(S): Confusing "medial right apex" with "medial left and right apex" in a lung cancer case changes the interpretation from a unilateral lesion to bilateral involvement. This could result in a staging upshift (e.g., from M0 to M1a~\citep{ramiporta2024tnm}), causing clinicians to abandon curative options (e.g., surgery or SBRT) and turn to systemic therapies or additional diagnostics.
Insignificant(I): Omitting "right" in "blunting of the costophrenic angle on the right" is often tolerable. Clinicians can easily identify the affected side through imaging comparison, and this does not typically alter clinical decisions.

\textbf{Disease severity} often guides treatment choices. Misreporting severity can lead to suboptimal decisions. For instance, choosing surgery when medication suffices, or delaying surgery when urgently needed.
S: The original report described focal consolidation with stable cardiomegaly. The modified version added "diffuse severe opacities bilaterally," falsely suggesting acute deterioration (e.g., ARDS or severe pneumonia). This may prompt unnecessary escalation of treatment (e.g., broad-spectrum antibiotics, aggressive ventilator settings); trigger additional diagnostics (e.g., bronchoscopy, BAL)~\citep{qadir2024ards}
I: Adding "mild elevation of the right hemidiaphragm" to a report on small bilateral effusions usually has no clinical consequence. This finding is common and non-specific, often due to benign factors like body position, mild diaphragm laxity, or small effusions. It rarely requires intervention and does not alter management.

\textbf{Lesion morphology} (e.g., margin, internal texture, and structural characteristics) is essential for tumor diagnosis and staging. Errors in description can lead to misclassification of malignancy and inappropriate treatment.
S: Changing "irregularly marginated" and "has grown" to "smoothly marginated" and "no change" understates malignant risk~\citep{wood2015nccn}. This may mislead clinicians into assuming the lesion is benign, resulting in: delayed diagnostic workup (e.g., CT, functional evaluation); missed early treatment opportunities; potential progression of undiagnosed cancer.
I: Replacing "irregularly marginated" with "poorly defined mass with uneven edges" conveys a similar clinical implication—both suggest malignancy and warrant further evaluation. Thus, such phrasing differences do not affect clinical interpretation or decision-making.

\textbf{Negative findings} are essential for ruling out differential diagnoses and avoiding unnecessary interventions. Misreporting such information can lead to overtreatment and increased medical burden.
S: Changing "no pneumothoraces" to "right pneumothorax" introduces a critical false positive. In patients with heart failure, this may trigger emergency responses such as chest tube placement, delay proper heart failure management, and risk unnecessary invasive procedures~\citep{roberts2023bts}.
I: Omitting mention of food content in the esophagus has little clinical impact. Such findings are common and do not typically influence diagnostic or therapeutic decisions.

Different diseases require appropriate \textbf{imaging modalities}. Incorrect statements about what can be seen on a given modality, or misleading follow-up recommendations, may result in wasted resources and misinformed clinical decisions.
S: Claiming that "esophageal mural thickening is clearly delineated on X-ray" is incorrect—such findings require CT~\citep{cha2021esophageal}. This may cause confusion, unnecessary concern, and inappropriate reliance on suboptimal diagnostic imaging.
I: Referring to "CT" without specifying "chest CT" is generally acceptable. Physicians can interpret the intent correctly based on prior reports and standard diagnostic pathways in clinical contexts.

\textbf{Quantitative descriptors}~\citep{heiman2025factchexcker}, such as lesion size or device position, are critical for diagnosis, staging, and treatment planning. Misstatements may lead to overtreatment, undertreatment, or delays in care.
S: Describing a 3-cm mass as a "very large mass" may falsely suggest a tumor $\ge$ 5 cm, resulting in higher T-stage classification under lung cancer TNM criteria~\citep{ramiporta2024tnm}. This can lead clinicians to: abandon curative surgery due to perceived inoperability; overestimate disease aggressiveness; order unnecessary invasive procedures or specialist referrals.
I: Adding a normal cardiothoracic ratio (e.g., 0.49) to a report about rib fractures has minimal clinical relevance. It does not impact fracture management or related clinical decisions.

\textbf{Temporal comparisons} are key to evaluating disease progression and treatment response. In conditions like hepatocellular carcinoma (HCC), small changes in lesion size over time can redefine response categories (e.g., partial response vs. progression)~\citep{vogel2025hcc}, directly influencing clinical decisions.
S: Changing the interpretation from "pulmonary edema improved" to "worsened" misleads clinicians into believing the patient is deteriorating. This may lead to: escalation of medication (e.g., higher diuretic doses); unnecessary ICU monitoring or imaging; increased healthcare costs and patient anxiety.
I: Replacing "less severe" with "slightly improved" conveys the same clinical direction (i.e., improvement). It does not alter risk assessment or treatment planning.

\textbf{Major internal contradictions} can undermine the credibility of a radiology report and may render the findings clinically unreliable. Such errors are often flagged in quality control and may directly endanger patient safety.
S: Stating "lungs are clear" immediately after describing large pleural effusion, atelectasis, and possible consolidation introduces a severe inconsistency. This may cause clinicians to question the report’s validity and hesitate to act on its findings.
I: Saying "no clear current evidence of chronic pulmonary changes" after suggesting chronic changes introduces a mild inconsistency, but not a true contradiction. Clinicians can still interpret the statement within clinical contexts and proceed appropriately.

\textbf{Expressions of uncertainty} in radiology reports guide clinicians toward cautious decision-making, including further testing or observation. Replacing uncertain language with unjustified certainty can mislead treatment and compromise patient safety.
S: Changing "may represent atelectasis or pneumonia" to a definitive "represents atelectasis," and describing possible free air as confirmed, may lead clinicians to: dismiss infection unnecessarily (e.g., withholding antibiotics); initiate premature surgical interventions based on presumed pneumoperitoneum. This compromises diagnostic objectivity and risks inappropriate treatment.
I: Phrases like "may be" vs. "appears to be" both reflect clinical uncertainty and do not meaningfully alter diagnostic interpretation or next steps.

Accurate use of \textbf{medical terminology} is essential for precise communication and diagnostic clarity. Substituting standard terms with vague or non-professional expressions can obscure clinical meaning and delay appropriate care.
S: Replacing "pleural effusion" with "pleural empyema" constitutes a critical error. This falsely suggests a localized infection requiring urgent drainage (e.g., chest tube placement), potentially leading to unnecessary invasive procedures, prolonged hospital stays, and inappropriate antibiotic use, while the actual cause (e.g., heart failure, malignancy) is overlooked. Similarly, substituting "atelectasis" with a phrase like "possible mass" incorrectly raises suspicion for malignancy, potentially triggering unnecessary biopsies, CT scans, and significant patient anxiety.
I: Using lay terms like "breathing tube" for "endotracheal tube" or "feeding tube" for "enteric tube" does not affect clinical interpretation when tube position and anatomy are described clearly. These variations preserve the report’s medical accuracy.

\textbf{Minor linguistic errors}, such as typos or grammatical mistakes, are common in clinical reports and typically do not affect interpretation. However, when noise alters the meaning of a sentence, it can mislead clinical decisions.
S: Changing "most likely due to low lung volumes and positioning" to "unlikely the cause" reverses the interpretation of a key finding. This may prompt unnecessary concern over a mediastinal abnormality, leading to further testing or referrals.
I: Typos like "lingulas", "growed", "studys", "atelectasi", "adenocarcinomia" are linguistically incorrect but do not affect the core diagnostic message. Clinicians can readily infer the intended meaning without clinical misunderstanding.

\textbf{Stylistic differences}, when semantically equivalent, usually do not impact clinical decisions. However, poor phrasing, ambiguous emphasis, or incorrect wording can reduce report clarity or lead to clinical misjudgment.
S: Changing a report that indicates clinical improvement (e.g., improving edema, resolving effusions, low lung volumes) to one that suggests deterioration (e.g., no edema, persistent effusions, normal lung inflation) reverses the overall interpretation. These conflicting signals may mislead treatment evaluation and disrupt appropriate care.
I: Changes in sentence order or phrasing using a different template do not alter diagnostic content. Both versions communicate the same findings and support the same clinical interpretation.

\section{Appendix: Additional Clarification}
\label{sec:appendix b}
CheXbert-F1 classifies each of 14 thoracic conditions into four categories: Positive, Negative, Uncertain, and Blank. The final score is computed as the micro-averaged F1 across all condition–label pairs. While this offers more granularity than binary classification, the metric still relies on discrete label matching and is limited in several ways: \textit{(1)} it cannot detect clinically important nuances such as changes in severity (e.g., "small effusion" vs. "large effusion"), \textit{(2)} it ignores how multiple findings interact or contradict each other (e.g., stating "no pleural effusion" in one sentence and "moderate right pleural effusion" in another), and \textit{(3)} it is insensitive to paraphrasing, hedging, or indirect language that may shift the clinical implication without altering the label category. Thus, while CheXbert is valuable for structured disease extraction, it lacks the semantic depth required to evaluate subtle but clinically significant variations in report generation. This concern has also been echoed in recent work—the GEMA score~\citep{zhang2025gema}.

We did not include certain metrics in our Meta-Evaluation framework for the following reasons. CheXprompt~\citep{zambrano2025smallmm} and RadCliQ~\citep{yu2023cxrreport} output structured error counts, such as per ReXVal category, rather than a scalar score, which makes them incompatible with our pairwise evaluation. As previously discussed, simple error counting is also inherently limited. Similarly, FineRadScore~\citep{huang2024fineradscore} employs an LLM to classify and explain errors line by line, but it does not provide a single numerical output and is highly dependent on sentence-level formulations. 

Furthermore, since our goal is not to rank metrics in an absolute sense, we chose not to conduct statistical significance tests such as t-tests. Instead, we aim to demonstrate that many existing metrics lack the ability to capture clinical semantics and to provide guidance for future metric design.

\textbf{Data Availability.} Our dataset is derived from MIMIC-CXR and ReXVal, both of which are distributed under the PhysioNet Credentialed Health Data License 1.5.0. Our work uses only these publicly available, credentialed datasets, and our derived dataset does not involve new data collection from patients. 
As a result, we cannot openly redistribute the report texts. Instead, we provide annotation guidelines, error taxonomy, and processing scripts so that credentialed users can reproduce our dataset from their own copies of MIMIC-CXR and ReXVal.

We have also initiated the process of submitting our derived dataset to PhysioNet for controlled release under the same license, ensuring compliance with patient privacy and reproducibility standards. Updated information can be found at~\url{https://github.com/ruochenli99/ReEvalMed}.

\textbf{Intended Use.} Our use of MIMIC-CXR and ReXVal strictly followed their intended purpose of research-only use. The derived dataset inherits the same restrictions and is provided solely for research, not for clinical decision-making or commercial applications.

\textbf{Privacy and Safety.} MIMIC-CXR and ReXVal have been de-identified by the dataset providers in compliance with HIPAA. No personally identifying information or offensive content is present in these datasets. Our derived dataset contains only de-identified report texts and annotation labels, and does not introduce any additional personally identifying or offensive content.

\textbf{Documentation.} All evaluation metrics used in this work were implemented via publicly available repositories, each with its own documentation and license (e.g., MIT, Apache 2.0). We used the official or widely adopted implementations without modification to ensure reproducibility, and we provide references to the original papers. 

MIMIC-CXR and ReXVal, which consist of English radiology reports (findings sections) from chest X-rays collected at a large U.S. academic medical center. Our derived dataset focuses on the findings of the report. No demographic attributes of patients are included.

\textbf{Experiment Details.} For generating rewritten reports, we used the DeepSeek-R1 7B model with fixed prompts and standard decoding settings, run locally on an NVIDIA H100 80GB GPU. Nonetheless, regardless of the model used, all generated modifications were carefully reviewed and validated by clinical experts to ensure correctness and clinical plausibility. The evaluation metrics were used via their publicly released implementations without modification, relying on their default model sizes and parameters. All experiments are lightweight and inference-only.

\textbf{Descriptive Statistics.} We report confidence intervals for each evaluation metric. 
For the Discriminative Score and Robustness Score, results are computed as the mean over 10 samples per evaluation aspect.

% \textbf{Data Consent.} Annotations were carried out by members of the research team with relevant domain expertise, who are not listed as co-authors of this paper. 
% No external annotators were recruited, and no crowd-sourcing platforms were used; hence, payment considerations are not applicable. As a result, no additional individual patient consent was required. 

\textbf{Use of AI Assistants.} ChatGPT and Grammarly were used to support writing and editing tasks, including drafting LaTeX tables, formatting references, and suggesting wording for writing paper.

\section{Appendix: Future Work}
\label{sec:appendix c}
Prior evaluation assessments, often based on a single correlation coefficient, may obscure important limitations. 
Our proposed aspect-based Meta-Evaluation framework aims to explicitly test whether metrics can distinguish between clinically significant and insignificant errors across a wide range of real-world report variations, offering a path toward more clinically aligned metric design.

Moving forward, we hope future metric development can incorporate these clinical aspects more explicitly. For example:

\textbf{Knowledge infusion} (e.g., integrating domain-specific ontologies or structured clinical guidelines) may help metrics reason about subtle but clinically meaningful variations.

\textbf{Chain-of-thought prompting or step-by-step reasoning} could guide LLM-based metrics to better assess the semantic consistency and clinical implications of generated content.

\textbf{Agent-based debate or multi-agent deliberation} may offer a way to simulate clinical decision-making dynamics when evaluating borderline cases or conflicting evidence.

An automatic and scalable workflow is critical, as high-quality dataset construction in clinical domains is inherently labor-intensive due to the need for expert validation. However, we designed our annotation protocol with future scalability in mind. Specifically, we found that LLM-based rewriting (guided by structured prompts) can generate clinically realistic error types across dimensions (e.g., severity, location, description). Current generations are generally acceptable to clinicians upon review. They significantly reduce clinical workload, especially when paired with targeted expert review rather than full manual rewriting. In future iterations, we plan to explore more robust semi-automated pipelines combining LLM generation and selective expert validation. This can enable scalable benchmark extension without sacrificing quality.

% This is an appendix.

\end{document}